\documentclass[10pt,journal,compsoc]{IEEEtran}
\ifCLASSOPTIONcompsoc
  \usepackage[nocompress]{cite}
\else
  \usepackage{cite}
\fi
\usepackage{cite}
\usepackage{tabularx}
\usepackage{caption}
\usepackage{physics}
\usepackage{algorithm2e}
\usepackage{amsmath,amssymb,amsfonts}
\usepackage{algorithmic}
\usepackage{graphicx}

\graphicspath{{./figures/}}
\usepackage{textcomp}
\usepackage{float}
\usepackage{wrapfig}
\usepackage{booktabs}
\usepackage{subcaption}

\usepackage{chemformula}
\usepackage{amsthm,amsmath,amssymb}
\usepackage{mathrsfs}
\usepackage{xcolor}

\ifCLASSINFOpdf
\else
\fi
\begin{document}
\title{Essential Number of Principal Components and Nearly Training-Free Model for Spectral Analysis}

\author{Yifeng Bie, Shuai You, Xinrui Li, Xuekui Zhang* and Tao Lu*
\thanks{The research was funded by Natural Sciences and Engineering Research Council of Canada (NSERC) Discovery (Grant No. RGPIN-2020-05938) (YB and TL), and Defense Threat Reduction Agency (DTRA) Thrust Area 7, Topic G18 (Grant No. GRANT12500317) (YB and TL),  Canada Research Chair (Grant No. 950231363) (SY and XZ).  This research was enabled in part by support provided by WestGrid (www.westgrid.ca) and Compute Canada (www.computecanada.ca). }
\thanks{Dr. Xuekui Zhang is with
the Mathematics and Statistics Department, University of Victoria, Victoria, BC, V8P 5C2, Canada (e-mail: Xuekui@UVic.ca).}
\thanks{Dr. Tao Lu is with the Department of Electrical and Computer Engineering, University of Victoria, Victoria, BC, V8P 5C2, Canada (e-mail: taolu@uvic.ca).}
\thanks{* Joint senior authors and corresponding authors.}
}

%
%

\markboth{arXiv,~Vol.~xx, No.~xx, January~2023}%
{Bie \MakeLowercase{\textit{et al.}}: Essential Number of Principal Components and Nearly Training-Free Model for Spectral Analysis}
%



\IEEEtitleabstractindextext{%
\begin{abstract}
Through a study of multi-gas mixture datasets, we show that in multi-component spectral analysis, the number of functional or non-functional principal components required to retain the essential information is the same as the number of independent constituents in the mixture set. Due to the mutual in-dependency among different gas molecules, near one-to-one projection from the principal component to the mixture constituent can be established, leading to a significant simplification of spectral quantification. Further, with the knowledge of the molar extinction coefficients of each constituent, a complete principal component set can be extracted from the coefficients directly, and few to none training samples are required for the learning model.  Compared to other approaches, the proposed methods provide fast and accurate spectral quantification solutions with a small memory size needed.
\end{abstract}

}

\maketitle

\IEEEdisplaynontitleabstractindextext

%
\IEEEpeerreviewmaketitle

\IEEEraisesectionheading{\section{Introduction}}
\label{sec:introduction}

\IEEEPARstart{B}{eing} capable of probing electronic and vibrational/rotational states, optical absorption spectroscopy is a proven tool for molecular quantification and classification with high sensitivity, low detection limit, and immunity to electromagnetic noises. Spectra are sensitive to small analyte variations and are often used to identify and quantify a sample's constituents. A plural of  spectroscopic analysis applications have been found in environmental monitoring~\cite{horz2004ammonia}, food industrial regulatory system~\cite{classen2017spectroscopic}, structural analysis~\cite{manso2009application, kumirska2010application}, and gas emission detection~\cite{oppenheimer2008probing}, etc. According to Beer-Lambert Law~\cite{prasad_biophotonics}, the absorbance of an analyte is proportional to its concentration at the same wavelength and within the same optical path length. Thus, a calibration curve measured from the standard samples can be used not only to determine the analyte but also to quantify it~\cite{pandhija2014calibration}. In the past, partial least squares regression (PLSR) is commonly used for spectral analysis due to its proven capacity for multivariate data analysis~\cite{doi:10.1021/ac00162a020,doi:10.1021/ac00162a021,Lin_Applications,Li_Rapid,Salehi_Aritficial,Lin_Applications,mehmood2012review,jiang2002wavelength}. However, novel approaches using machine learning for sample classification~\cite{luyun} and quantification~\cite{Yang_Comparison,borin2006least} have shown improved performance. With sufficient training data, machine learning ~\cite{patterson2010letting,goh2017deep,tarca2007machine,shalev2014understanding} can project samples from spectral space to constituent concentration space. Additionally, if the predictions are to be made from indirect factors, such as gas constituents in a mixture being highly correlated to each other, machine learning will outperform conventional methods~\cite{luyun,liu2015determination}.

In field applications such as on-site toxic gas detection and mineral analysis, memory size, processing speed and power consumption are limiting factors. Therefore, data pre-processing methods are often employed before the machine learning model to reduce the input data dimension. Principal components analysis (PCA)\footnote{In this article, to avoid confusion with its functional counterpart, we also call PCA as non-functional principal component analysis.} is one of the most popular methods for dimension reduction~\cite{thomaz2010new,croux2002sign}. PCA, which reduces the data dimension by projecting it onto a set of orthogonal bases, however, neglects the functional nature that many data possess. In contrast, functional principal components analysis (fPCA) is an extension of multivariate PCA into the functional case to reduce the infinite dimension of a functional predictor and to explain its dependence structure by a reduced set of uncorrelated variables\cite{aguilera2013penalized}. By decomposing the functional data into a linear combination of functional principal components (fPCs), fPCA is more efficient than PCA in extracting core information from functional data, which suits the nature of spectral analysis~\cite{nie2020sparse,yao2005functional,dona2009application,shang2014survey}.

With the significantly reduced data dimension through preprocessing, highly efficient machine learning models are normally used for spectral analysis with lower power consumption and faster processing time, making real-time spectral analysis in field applications feasible. One of the suitable candidates is Extreme Gradient Boosting (XGBoost)~\cite{xgboost}, which is a supervised machine learning algorithm being used in a variety of applications such as medical~\cite{torlay2017machine,wang2019irespy}, image recognition~\cite{dong2018gaofen}, fault detection~\cite{chakraborty2019early}, etc. XGBoost has been empirically proven to be fast with a high accuracy that can even outperform deep learning~\cite{nielsen2016tree}. Compared with other deep learning methods, such as feed-forward neural networks~\cite{sazli2006brief} or convolutional neural networks~\cite{o2015introduction}, XGBoost is a  simpler model based on the regression tree. The model requires fewer parameters to tune and is faster than deep learning. In this article, our original goal was to use XGBoost to quantify sample constituents from its absorption spectrum. We discovered, however, it is sufficient to use PCA or fPCA in combination with linear regression models for the task directly, making the addition of any follow-up machine learning models a redundancy.

In this  article, we first show that for a set of mixtures with $K$ independent constituents, the dimension of spectra data can be significantly reduced through either functional or non-functional principal components analysis. In addition, keeping the same number of top $K$ principal components is sufficient for spectral quantification. By formulating a linear regression model (LR), we demonstrate that a simple LR with $K$ principal components outperforms both PLSR and XGBoost. Identifying the near one-on-one projection from the gas constituent to the principal component, we further simplify the LR model for higher efficiency and less memory usage. In the final stage, by extracting principal components directly from gas molecular extinction coefficients, we formulate an unsupervised spectral quantification model that requires orders of magnitude fewer training samples. The model not only provides accurate quantification of gas constituents but also provides a precise estimation of the overall detection system noises. In this article, we focus our discussions on fPCA and refer PC to the principal component of both functional and non-functional ones for convenience. However, as discussed in Supplementary Information Section S.1, all conclusions of fPCA in this article can be directly extended to the non-functional PCA.

\begin{figure}[thbp]
    \centering
    \includegraphics[width=1\linewidth]{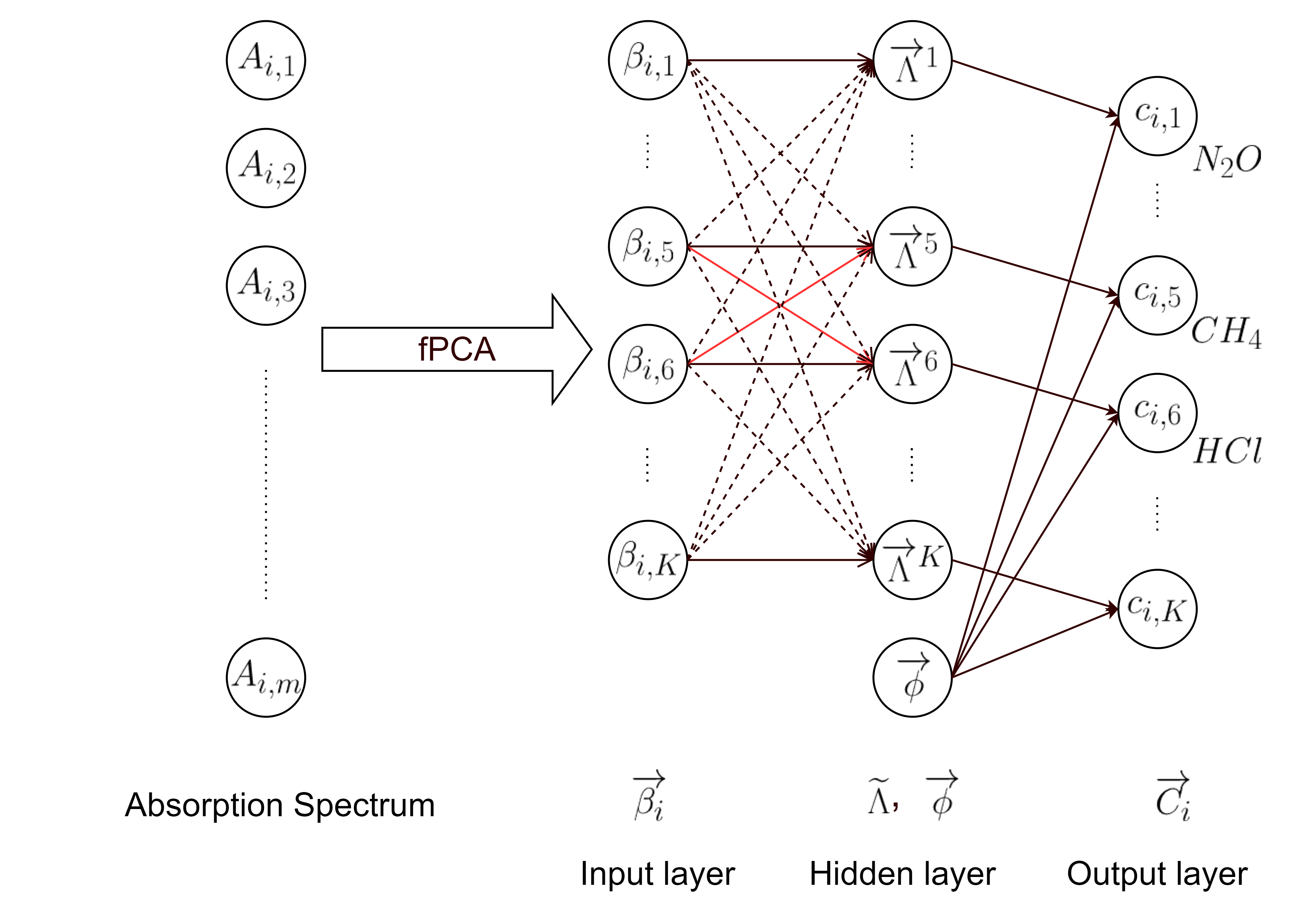}
    \caption{Determination of ${\tilde \Lambda}$ through a single-layer linear regression model.\label{fig:MLP}}
\end{figure}

\section{dataset}\label{Sec:Dataset}

To verify our models, we generated three groups of 9-gas mixture mid-infrared absorption spectroscopy datasets following the method described in~\cite{luyun}.  The datasets, which are available at~\cite{sensornet}, contain random mixtures of \ch{C_2H_6}, \ch{CH_4}, \ch{CO}, \ch{H_2O}, \ch{HBr}, \ch{HCl}, \ch{HF}, \ch{N_2O} and \ch{NO} with different mixing ratios. In particular, the first group of datasets (Group I) contain constituent concentrations uniformly distributed between 0 and 10$~{\mu}$M.  The constituent concentrations of both the second (Group II) and third (Group III) groups of datasets are log distributed within the range of 100~pM-10~$\mu$M for Group II and 10~pM-1~mM for Group III. Further, to make Group II and III datasets compatible for spectral classification, each of the 9 gas constituents only appears in half of the mixture samples. In the measurement system, the mixture sample is placed between a light source and a spectrometer. The optical path length of the sample is defined as $b$ in the unit of $cm$. To mimic real ambient environments,  zero mean with a standard deviation $\sigma$ Gaussian relative intensity noises (RIN) $\rho(\lambda)={\cal N}(0,\sigma)$ are added to the light source intensities $I_0(\lambda)$ such that $I_0(\lambda)={\bar I_0}(1+\rho(\lambda))$.  Here, for simplicity, we assume the average intensity of the light source, ${\bar I_0}$, is wavelength independent. The signal-to-noise ratio, $SNR=-10log_{10}\sigma$, is defined as the ratio between the light source intensity to the noise intensity in the unit of dB. In our model, we only consider the noises from the light source and ignore all other noises such as those from the spectrometer while those noises can be easily included as part of $\rho$. The absorption spectrum $\ket{A(\lambda)}$ of a sample is defined as\footnote{In the original dataset, $\ket{A(\lambda)}$ is defined in dB, which is equivalent to Eq.~\eqref{eq:absorbance} by setting $b$ 10 times larger.}
\begin{equation}
    \ket{A(\lambda)}=-log_{10}\frac{I(\lambda)}{{\bar I_0}}
    \label{eq:absorbance}
\end{equation}
where $I(\lambda)$ is the light intensity after passing through the sample, and can be computed through the Bell-Lambert law~\cite{prasad_biophotonics}. In each group, we generated four 100,000-sample datasets with SNRs of 10~dB, 20~dB, 30~dB, and 40~dB respectively.
\section{Methodology}
\subsection{Minimum number of  principal components}
\label{sec:fpca_pca}
\begin{figure}[thbp]
    \centering
    \includegraphics[width=0.9\linewidth]{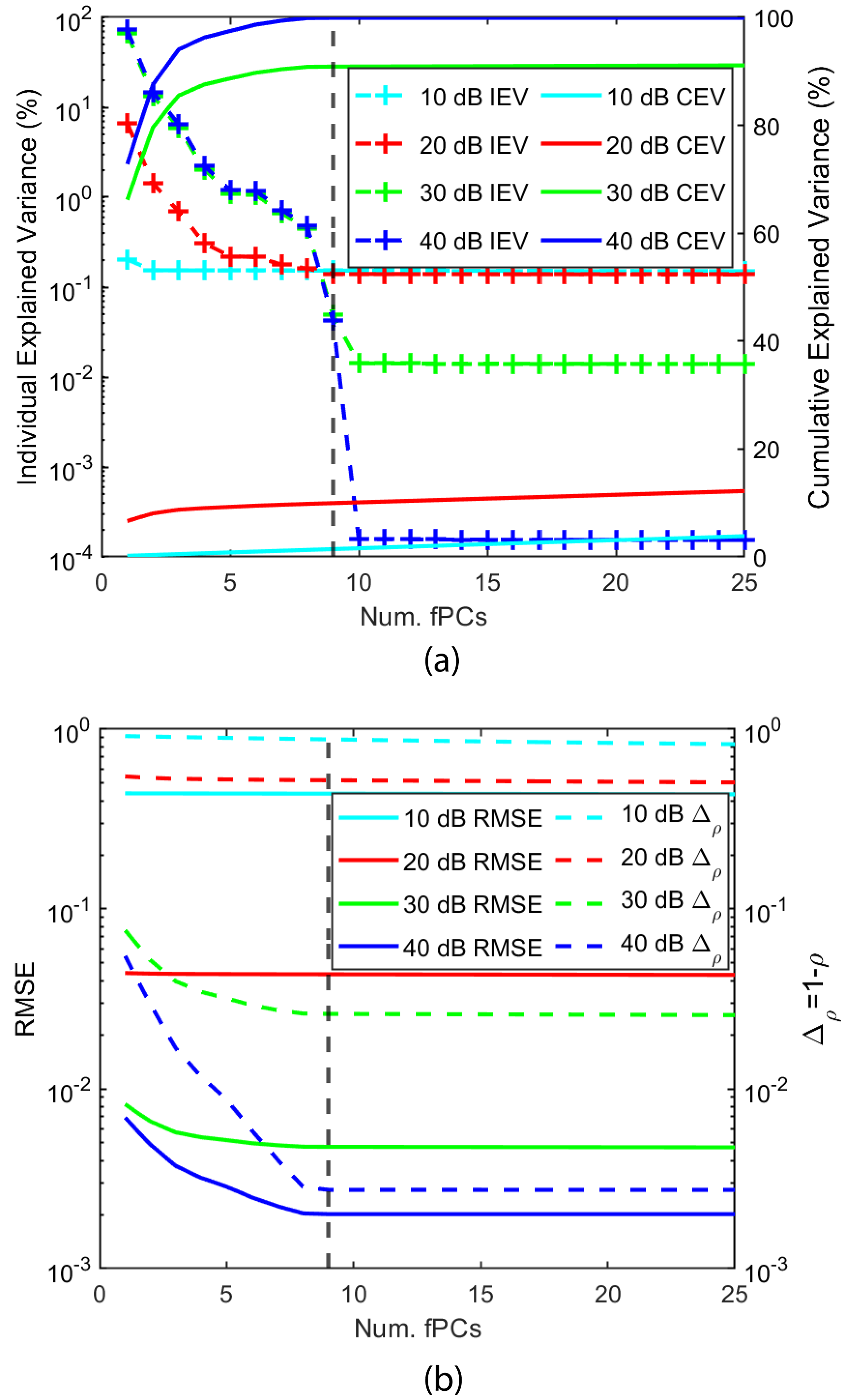}
    \caption{(a) The individual explained variance (IEV, dashed lines to the left y-axis) and cumulative explained variance (CEV, solid lines to the right y-axis) vs. the number of principal components.  (b) The average RMSE (solid lines to the left y-axis) and the residual correlation coefficient $\Delta_\rho=1-\rho$ (dashed lines to the right y-axis) of the Group I training datasets between the original and reconstructed spectrum as a function of the number of PCs adopted for reconstruction. The SNRs of the corresponding datasets are 10~dB (cyan), 20~dB (red), 30~dB (green), and 40~dB (blue) respectively. The black dashed line is positioned at the 9th top principal component.\label{fig:min_no_pc}}
\end{figure}

It is well known that PCA or fPCA are frequently used in data pre-processing for data dimension reduction. In most scenes, the top few components that explain the most variance are sufficient to reconstruct the data without significant distortion. In the task for spectral quantification, the question is how many top principal components we should retain. According to the Bell-Lambert law~\cite{prasad_biophotonics}, at a wavelength $\lambda$, the absorption spectrum $\ket{A(\lambda)}$ of a $K$-constituent mixture can be written as a linear superposition of the molar extinction coefficients of all constituents $\{\ket{\epsilon_k(\lambda)}=\norm{\epsilon_k}\ket{{\hat \epsilon}_k(\lambda)}, k=1,\ldots, K\}$,
\begin{equation}
    \ket{A(\lambda)}=\sum_{k=1}^Kbc_k\norm{\epsilon_k}\ket{{\hat \epsilon}_k(\lambda)}+\ket{n(\lambda)}\label{Eq:absorbance}
\end{equation}
where $c_k$ is the molar concentration of the $k$-th constituent in the unit of mole per liter $(M/L)$ and $\ket{n(\lambda)}$ is the noise spectrum.  $\ket{{\hat \epsilon}_k(\lambda)}$ is the unitless normalized molar extinction coefficients such that $\bra{{\hat \epsilon}_k}\ket{{\hat \epsilon}_k}=1$ and $\norm{\epsilon_k}=\sqrt{\bra{\epsilon_k}\ket{\epsilon_k}}$ is its magnitude in the unit of $({M}^{-1}L\cdot{cm}^{-1})$. Although $\ket{A(\lambda)}$ resides in an infinite dimensional Hilbert space due to noises, $c_k$ are the projection coefficients to the mutually independent functional set $\{\ket{\epsilon_k(\lambda)}\}$ in a $K$-dimension sub-space. In the cases when the noise power is small, the top $K$ principal components form the basis set for the $K$-dimension sub-space since signal presides. On the other extreme, when the noises dominate the spectrum, then signals are submerged under noise, and we are unable to find the subspace to obtain $c_k$ regardless of the number of top principal components used. Nevertheless, top principal components from 1 to $k$ are only required, as a few additional components do not enhance quantification.
\begin{figure*}[htb]
    \centering
 \includegraphics[width=1\linewidth]{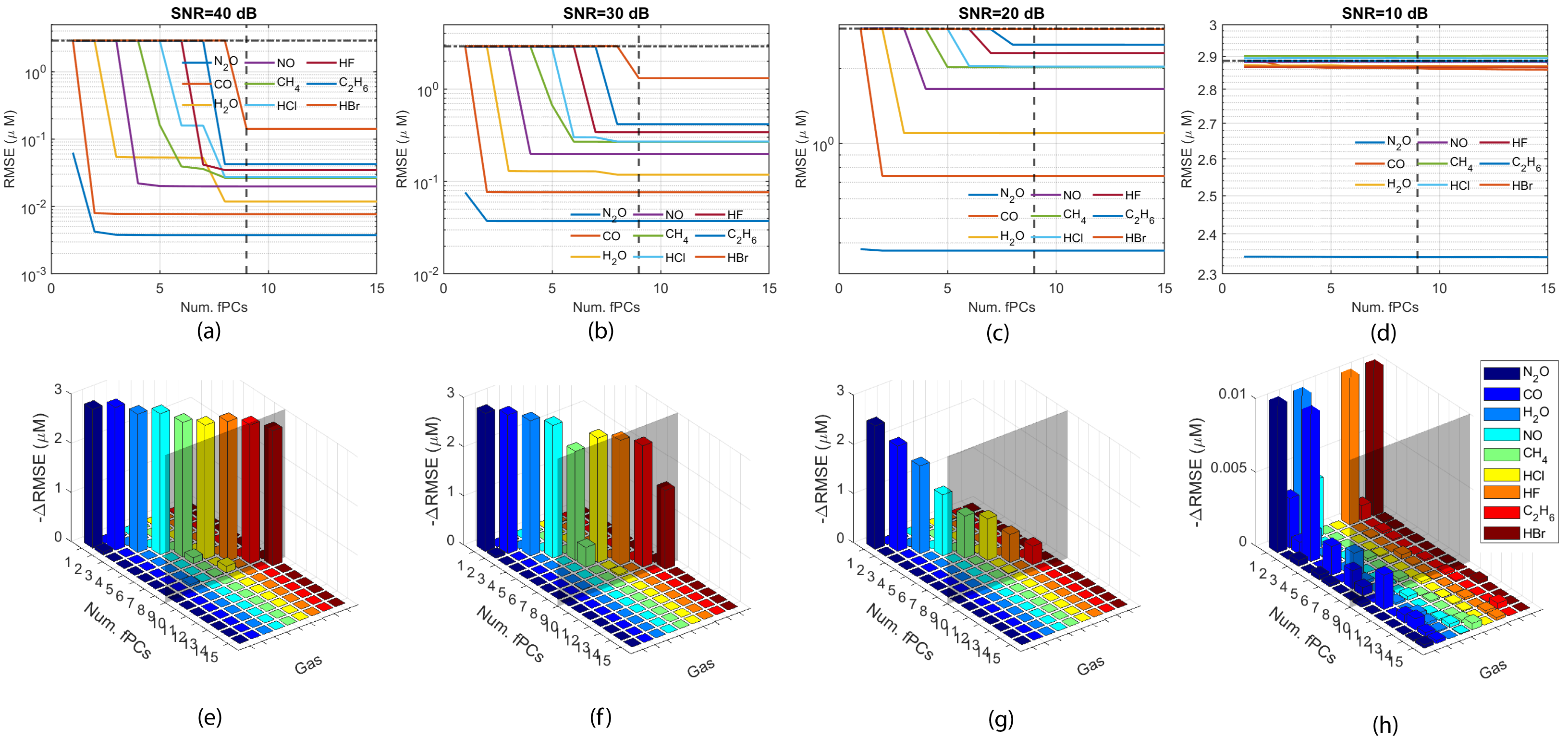}
    \caption{\label{fig:prediction_rmse_vs_fpc} RMSE of spectral quantification vs. the number of fPCs adopted and gas types using linear regression with fPCA (fPCA-LR). The SNRs are (a) 40~dB, (b) 30~dB, (c) 20~dB and (d) 10~dB. The horizontal dashed line at $2.887~{\mu}M$ represents the RMSE of random guesses. The vertical dashed line is placed at the 9th fPC. The RMSE decrement $(-\Delta{RMSE})$ vs. the number of fPCs for each gas are plotted in (e) to (h) when SNRs are (e) 40~dB, (f) 30~dB, (g) 20~dB and (h) 10~dB. The vertical grey planes are placed at the 9th fPC.}
\end{figure*}

\subsection{Spectral quantification with linear regression}
\begin{figure*}[htb]
    \centering
 \includegraphics[width=1\linewidth]{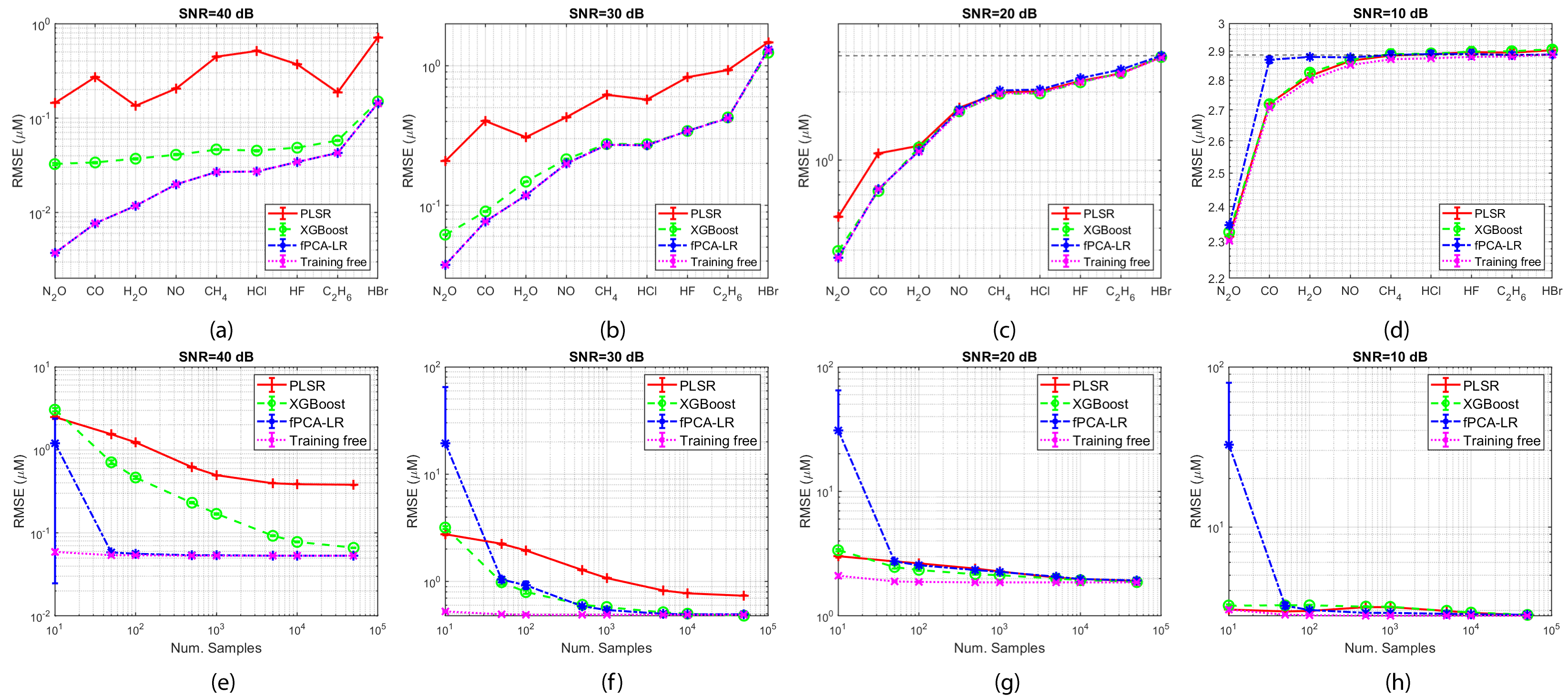}
    \caption{\label{fig:RMSE_all_vs_SNR_result} Average RMSE of spectral quantification using PLSR (red solid line with plus markers), XGBoost (green dashed line with circle markers), Linear regression with fPCA (fPCA-LR, blue dash-dotted line with star markers), direct quantification using fPCA (magenta dotted line with cross markers) and TF (cyan solid line with diamond markers) methods. The SNRs are (a) 40~dB, (b) 30~dB, (c) 20~dB and (d) 10~dB. The black dashed line at $2.887 \ {\mu}M$ represents the RMSE of random guesses. In (e) to (h), the total RMSE of all gases vs. the number of training samples adopted when the SNRs are (e) 40~dB, (f) 30~dB, (g) 20~dB, and (h) 10~dB respectively.}
\end{figure*}

Observing the linear relation between $\ket{A(\lambda)}$ and $\ket{\epsilon_k(\lambda)}$, we start our discussions by deriving a linear regression model (FPCA-LR). Further, as shown in the later section, due to the nearly orthogonal nature of $\ket{\epsilon_k}$ among constituents, a nearly one-to-one correspondence can be found between  $\ket{\epsilon_k(\lambda)}$ and principal components. Utilizing this property, the model can be simplified to a direct quantification procedure using principal component scores. To start, we first apply fPCA to an $N$-sample training dataset and retain $K$ orthonormal functional principal components and mean $\{\ket{\phi_l(\lambda)},\ket{u(\lambda)}; l=1,\ldots,K\}$~\footnote{We here adopt $\ket{u(\lambda)}$ as part of the basis set. For some fPCA/PCA implementations that do not extract mean, the formulated model is equally valid by setting the mean to a zero function $\ket{u(\lambda)}\equiv\ket{0}$.}. The absorption spectrum $\{\ket{A_i(\lambda)},i=1,\ldots,N\}$ of the $i$-th sample that is measured at $M_\lambda$ wavelengths $\{\lambda=\lambda_j,j=1,\ldots,M_\lambda\}$ can be approximated as
\begin{equation}
    \ket{A_i(\lambda)}=\sum_{l=1}^K\beta_{l,i}\ket{\phi_l(\lambda)}+\ket{u(\lambda)}+\ket{r_i(\lambda)}.
\end{equation}
where $\ket{r_i(\lambda)}$ is the residual spectrum that represents the part of the original spectrum $\ket{A_i(\lambda)}$ that can not be approximated by the linear superposition of the $K$ selected principal components. Again, under the assumption that the SNR is sufficiently high, the magnitude of the residual spectrum is small. Substituting $\ket{A_i(\lambda)}$ in Eq.~\eqref{Eq:absorbance} and defining $c_{k,i}$ as the concentration of $k$-th constituent in sample $i$, $\ket{n_i(\lambda)}$ the noise spectrum of the same sample, we get
\begin{equation}
\begin{array}{l}
    \sum_{l=1}^K\beta_{l,i}\ket{\phi_l(\lambda)}+\ket{u(\lambda)}+\ket{r_i(\lambda)}\\
    =\sum_{k=1}^Kbc_{k,i}\norm{\epsilon_k}\ket{{\hat \epsilon}_k(\lambda)}+\ket{n_i(\lambda)}
    \end{array}
\end{equation}
The equation can be further simplified by applying $\bra{\phi_p(\lambda)}$ on both sides\footnote{Here, following the convention in bra-ket space, we define $\bra{g(\lambda)}\ket{f(\lambda)}=\sum_{i=1}^M g^*(\lambda_i)\cdot{f}(\lambda_i)$ as the inner product of two functions $f(\lambda)$ and $g(\lambda)$ with $g^*(\lambda)$ being the complex conjugate of the function $g(\lambda)$.} and adopting the orthonormal  relation $\bra{\phi_p(\lambda)}\ket{\phi_l(\lambda)}=\delta_{p,l}$ with $\delta_{p,l}$ being the Kronecker delta,
\begin{equation}
\begin{array}{ll}
\beta_{p,i}=&\sum_{k=1}^Kbc_{k,i}\norm{\epsilon_k}\bra{\phi_p(\lambda)}\ket{{\hat \epsilon}_k(\lambda)}\\
&+\bra{\phi_p(\lambda)}(\ket{n_i(\lambda)}-\ket{u(\lambda)}-\ket{r_i(\lambda)})\label{Eq:beta}
\end{array}
\end{equation}
Note that $\bra{\phi_p(\lambda)}\ket{r_i(\lambda)}\equiv 0$, applying Eq.~\eqref{Eq:beta} to all samples in the training data set for all principal components $p$, we have $K\times{N}$ linear equations which can be expressed into a matrix form
\begin{equation}\label{Eq_matrix_beta_to_c}
    {\tilde \beta}=(b{\tilde \psi}\cdot{\tilde \epsilon}){\tilde C}+({\tilde N}-{\tilde u})
\end{equation}
by defining $K\times{N}$ matrices as ${\tilde \beta}$, ${\tilde u}$ and ${\tilde N}$, a $K\times{K}$ overlap matrix as ${\tilde \psi}$ and a $K\times{N}$ concentration matrix as ${\tilde C}$ according to
\begin{equation}
    \begin{array}{lll}
    {\tilde \beta}&=&\begin{bmatrix}\beta_{1,1}&\ldots&\beta_{1,N}\\
    \vdots&\ddots&\vdots\\
    \beta_{K,1}&\ldots&\beta_{K,N}
    \end{bmatrix}\\
    \end{array}
    \end{equation}
    \begin{equation}
        \begin{array}{lll}

    {\tilde \psi}&=&\begin{bmatrix}\bra{\phi_1(\lambda)}\ket{{\hat \epsilon}_1(\lambda)}&\ldots&\bra{\phi_1(\lambda)}\ket{{\hat \epsilon}_K(\lambda)}\\
    \vdots&\ddots&\vdots\\
    \bra{\phi_K(\lambda)}\ket{{\hat \epsilon}_1(\lambda)}&\ldots&\bra{\phi_K(\lambda)}\ket{{\hat \epsilon}_K(\lambda)}
    \end{bmatrix}\\
    \end{array}
    \end{equation}
    \begin{equation}
        \begin{array}{lll}
   {\tilde C}&=&\begin{bmatrix}c_{1,1} & \ldots & c_{1,N}\\
    \vdots & \ddots & \vdots\\
    c_{K,1} & \ldots & c_{K,N}
    \end{bmatrix}\\
    \end{array}
    \end{equation}
    \begin{equation}
        \begin{array}{lll}
      {\tilde \epsilon}&=&\begin{bmatrix}\norm{\epsilon_1} &  & \\
     & \ddots & \\
     &  & \norm{\epsilon_K}
    \end{bmatrix}\\
    \end{array}
    \end{equation}
    \begin{equation}
        \begin{array}{lll}
   {\tilde u}&=&\begin{bmatrix}\bra{\phi_1(\lambda)}\ket{u(\lambda)}&\ldots&\bra{\phi_1(\lambda)}\ket{u(\lambda)}\\
       \vdots&\ddots&\vdots\\
  \bra{\phi_K(\lambda)}\ket{u(\lambda)}&\ldots&\bra{\phi_K(\lambda)}\ket{u(\lambda)}\\
  \end{bmatrix}\\
    \end{array}
    \end{equation}
    \begin{equation}
        \begin{array}{lll}
    {\tilde N}&=&\begin{bmatrix}\bra{\phi_1(\lambda)}\ket{n_1(\lambda)}&\ldots&\bra{\phi_1(\lambda)}\ket{n_N(\lambda)}\\
        \vdots&\ddots&\vdots\\
   \bra{\phi_K(\lambda)}\ket{n_1(\lambda)}&\ldots&\bra{\phi_K(\lambda)}\ket{n_N(\lambda)}
   \end{bmatrix}\\
    \end{array}
\end{equation}
 Note that values of ${\tilde \beta}$, ${\tilde u}$ and ${\tilde C}$ are known  by applying fPCA to the training dataset while $(b{\tilde \psi}\cdot{\tilde \epsilon})$ and ${\tilde N}$ are unknown. It is worth mentioning that there are only $K$ independent elements in ${\tilde u}$ as the values of elements in each row are identical. Further, assuming the noises of each sample are independent and identically distributed (i.i.d.), the expectation $E\{{\tilde N_p}\}=E\{\bra{\phi_p}\ket{n_1}\}=\ldots=E\{\bra{\phi_p}\ket{n_N}\}$ has only $K$ unknowns as well since the values in each row are identical.  In addition, we have $K\times{K}$ unknowns from $E\{b{\vec \psi^p}\}$. In total, there are $N\times{K}$ linear equations in Eq.~\eqref{Eq_matrix_beta_to_c} with $K^2+K$ unknowns. Therefore, when $N>{K+1}$, Eq.~\eqref{Eq_matrix_beta_to_c} represents an overdetermined  system and all unknowns form $E\{b{\tilde \psi}\}$ and $E\{{\tilde N}\}$ can be estimated in the least square sense. When SNR is sufficiently large, $\ket{n_i}\approx\rho(\lambda)={\cal N}(0,\sigma)$ with ${\cal N}(0,\sigma)$ being zero mean and $\sigma^2$ variance Gaussian distribution, and $E\{{\tilde N}\}={\tilde 0}$. At lower SNR, however, the approximation no longer holds since the noise term $\ket{n_i}=\log_{10}(1+\rho(\lambda))$. The noise expectation is non-zero but still can be evaluated following the procedure above.

The estimated overlap matrix $E\{(b{\tilde \psi})\}$ and noise vector $E\{{\vec N}\}=[E\{\bra{\phi_1(\lambda)}\ket{n_i(\lambda)}\},\ldots,E\{\bra{\phi_K(\lambda)}\ket{n_i(\lambda)}\}]^T$ can be used for sample spectral quantification by applying Eq.~\eqref{Eq_matrix_beta_to_c} with ${\tilde C}$ being the only unknowns to be solved. To fully utilize the power of machine learning, we can reformat Eq.~\eqref{Eq_matrix_beta_to_c} into the governing equations of  a single-layer linear regression model~\cite{montgomery2021introduction} as shown in Fig.~\ref{fig:MLP}
\begin{equation}
    {\vec C}_i = {\tilde \Lambda}{\vec \beta}_i+{\vec \kappa}\label{Eq:C_Lambda}
\end{equation}
where ${\vec C}_i=[c_{1,i},\ldots,c_{K,i}]^T$ is the concentration vector of sample $i$, and ${\vec \beta}_i=[\beta_{1,i},\ldots,\beta_{K,i}]^T$ is fPC score vectors of the sample. ${\tilde \Lambda} =E\{ {\tilde \epsilon}^{-1}(b{\tilde \psi})^{-1}\}$ is a $K\times{K}$ weight matrix of the model and ${\vec \kappa} = E\{{\tilde \epsilon}^{-1}(b{\tilde \psi})^{-1} ({\vec u}-{\vec N})\}$ becomes a $K\times{1}$ bias vector. To apply the linear regression model, we first extract the top $K$ fPCs from the training dataset. The set of fPCs, whose number is equal to that of the gas components in a mixture, will be adopted to obtain the fPC score vector ${\vec \beta_i}$ of each training sample $i$. During training, sample fPC scores are fed into the model to learn ${\tilde \Lambda}$ and ${\vec \kappa}$ such that the in-sample error between the model output ${\vec C_i}$ and the ground truth gas concentrations are minimized. During testing, testing sample fPC score ${\vec \beta}$ is first computed from the spectrum using the fPCs from training. The scores become the inputs of the trained linear regression model to predict the gas concentrations.

\subsection{Direct quantification with fPCs}
\begin{figure*}[thbp]
    \centering
    \includegraphics[width=1\linewidth]{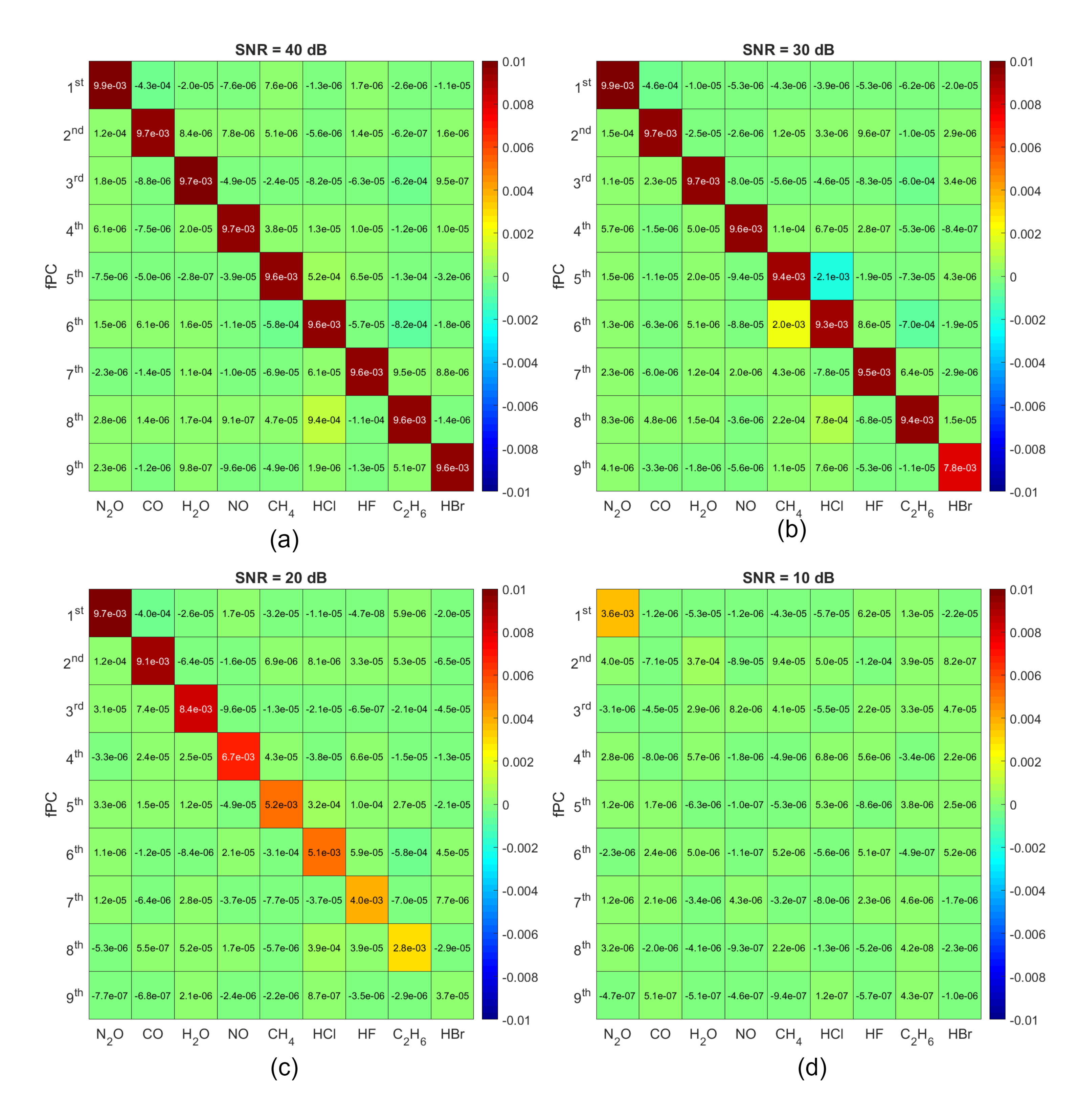}
\caption{Heat maps of the matrix $({\tilde \epsilon}\cdot\tilde{\Lambda})^T$ at the SNRs of (a) 40~dB, (b) 30~dB, (c) 20~dB and (d) 10~dB using Group I training datasets. \label{fig:Lambda_heat_map}}
\end{figure*}

Further studies in Section~\ref{sec:result} suggest that ${\tilde \psi}$ is highly diagonal. This is due to the mutual in-dependency of gas constituents, and consequently, only one or few fPCs are sufficient to explain each gas component. Using this property, spectral quantification can be greatly simplified. For example, if $\ket{\phi_p}$ has much stronger correlation to $\ket{\epsilon_p}$ than any other gas components, then by approximating $\{{\tilde \Lambda}_{p,k}\approx{0};k\ne{p}\}$ and following Eq.~\eqref{Eq:C_Lambda}, the training of $c_{i,p}$ can be reduced to a single linear equation
\begin{equation}
    c_{i,p}\approx\Lambda_{i,p}\beta_{i,p}+\kappa_p
\end{equation}
where both $\Lambda_{i,p}$ and $\kappa_p$ can be obtained by the linear least square fitting. On the other hand, if $\ket{\phi_p}$ has non-negligible overlaps to more than one gas component, Eq.~\eqref{Eq:C_Lambda} can still be significantly simplified by extracting only the equations that correspond to the gas components involved.
\begin{figure}[thbp]
    \centering
    \includegraphics[width=1\linewidth]{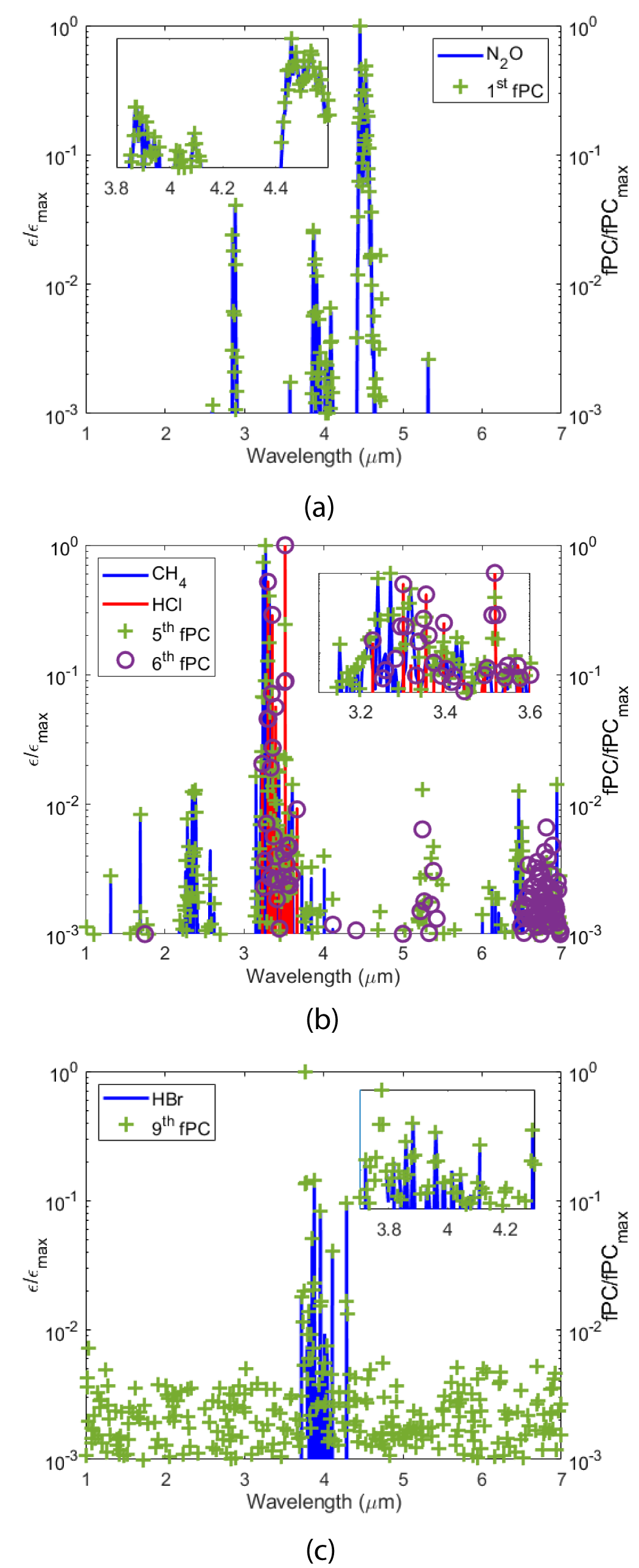}
\caption{The normalized molar extinction coefficients of (a) \ch{N_2O} (blue curve to the left axis) and 1st fPC (green plus markers to the right y-axis), (b) \ch{CH_4} (blue curve) and \ch{HCl} (black curve), both to the left y-axis, and 5th fPC (red plus markers) and 6th fPC (purple circle markers), both to the right y-axis) and (c) \ch{HBr} (blue curve) and 9th fPC (green plus markers). \label{fig:cross_section_fpc}}
\end{figure}

\begin{figure}[thbp]
    \centering
    \includegraphics[width=1\linewidth]{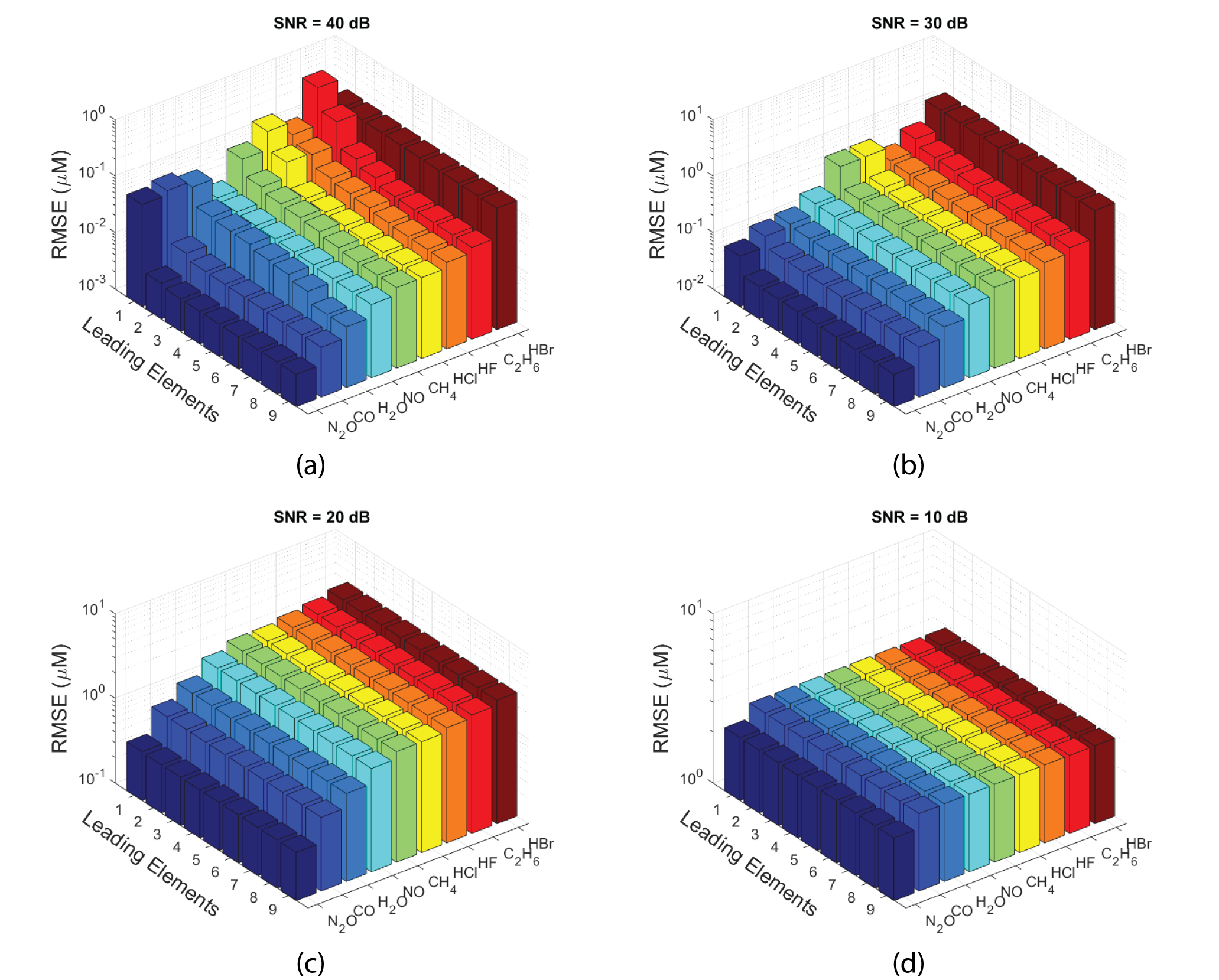}
\caption{Gas concentration prediction RMSE vs. the number of leading row elements in ${\tilde \Lambda}$ retained in direct quantification. The SNRs are (a) 40~dB, (b) 30~dB, (c) 20~dB and (d) 10~dB. \label{fig:direct_quantification}}
\end{figure}

\subsection{Nearly training free (TF) learning model}
In the above subsection, quantification relies on ${\vec N}$ and ${\tilde \psi}$, both of which can be obtained from training datasets. The training procedure does not require a priori knowledge of the molar extinction coefficients. However, the training process can be greatly simplified if $\ket{\epsilon_k(\lambda)}$ and $b$ are both known. In fact, a learning model can be directly established without a training dataset. Here, instead of obtaining the fPCs $\{\ket{\phi_l(\lambda)},\ket{u(\lambda)};l=1,\ldots,K\}$ from an $N\times{M_\lambda}$ training dataset, we obtain them from the $K\times{M_\lambda}$ molar extinction coefficient set $\{\ket{{\hat \epsilon}_k(\lambda)},k=1,\ldots,K\}$ through fPCA or PCA. In our case, $N=100,000$, $M_\lambda=1,000$, and $K=9$. Therefore, extracting fPCs from the $K$ molar extinction coefficients results in a more than $10^5$ fold increase in efficiency. The procedure is as follows, we first reconstruct the molar extinction coefficients from the fPC set
\begin{equation}
    \ket{\epsilon_k(\lambda)}=\sum_{l=1}^K\beta_{k,l}\norm{\epsilon_k}\ket{\phi_l(\lambda)}+\norm{\epsilon_k}\ket{u(\lambda)}\label{Eq:training_epsilon}
\end{equation}
Here, no residual spectrum $\ket{r(\lambda)}$ is required as the fPCs span the complete solution space for $\ket{\epsilon_k(\lambda)}$. Substituting Eq.~\eqref{Eq:training_epsilon} in Eq.\eqref{Eq:absorbance} to a testing spectrum $\ket{A(\lambda)}$ and applying $\bra{\phi_p(\lambda)}$ to both sides of Eq.~\eqref{Eq:training_epsilon}, we obtain
\begin{equation}
    \bra{\phi_p(\lambda)}\ket{A(\lambda)}=\sum_{k=1}^K\left[\beta_{k,p}+\bra{\phi_p}\ket{u(\lambda)}\right]bc_k\norm{\epsilon_k}+\bra{\phi_p(\lambda)}\ket{n(\lambda)}\label{Eq:phi_p}
\end{equation}
Eq.~\eqref{Eq:phi_p} can again apply to all fPCs and be written in a matrix form
\begin{equation}
    {\vec A}=b{\tilde \beta^{\prime T}}\cdot{\tilde \epsilon}\cdot{\vec C}+{\vec N^\prime}\label{Eq:trainingfree}
\end{equation}
where
\begin{equation}
    {\vec A}=\left[\bra{\phi_1(\lambda)}\ket{A(\lambda)},\ldots,\bra{\phi_K(\lambda)}\ket{A(\lambda)}\right]^T\\
    \end{equation}
    \begin{equation}
    {\vec N^\prime}=\left[\bra{\phi_1(\lambda)}\ket{n(\lambda)},\ldots,\bra{\phi_K(\lambda)}\ket{n(\lambda)}\right]^T\\
    \end{equation}
    \begin{equation}
    {\tilde \beta^\prime}=\begin{bmatrix}\beta_{1,1}+\bra{\phi_1}\ket{u(\lambda)} & \ldots & \beta_{1,K}+\bra{\phi_K}\ket{u(\lambda)}\\
    \vdots &\ \ddots & \vdots\\
    \beta_{K,1}+\bra{\phi_1}\ket{u(\lambda)} & \ldots & \beta_{K,K}+\bra{\phi_K}\ket{u(\lambda)}\\
    \end{bmatrix}
\end{equation}
Note that all numerical values of $\bra{\phi_p(\lambda)}\ket{A(\lambda)}$, $\beta_{k,p}$, $\bra{\phi_p}\ket{u(\lambda)}$ and $b$ are known. At high SNR, $E\{{\vec N^\prime}\}={\vec 0}$. In this case, no additional training is necessary, and Eq.~\eqref{Eq:trainingfree} represents $K$ linear equations with $K$ unknowns $E\{{\vec C}\}$, which can be solved directly.

In the case when $b$ is not a priori known and/or the SNR is low, then learning $b$ and a set of $K$ unknowns $E\{{\vec N^\prime}\}$ is necessary. However, the $K+1$ learning parameters are significantly fewer than the previous linear regression model and far fewer training samples are needed. In practice, a single-layer linear regression with $K$-input, $K$-output, and $K$-neuron is adopted to solve for $b$ and $N^\prime$. To obtain the gas concentrations, Eq.~\ref{Eq:trainingfree} is rewritten as:
\begin{equation}
    {\vec C} = \left(b{\tilde \beta^{\prime T}}\cdot{\tilde \epsilon}\right)^{-1}({\vec A} - {\vec N^\prime})\label{Eq:trainingfree_prediction}
\end{equation}
which can again be evaluated through the linear regression model expressed by Eq.~\eqref{Eq:C_Lambda} with ${\tilde \Lambda}=E\left\{ \left(b{\tilde \beta^{\prime T}}\cdot{\tilde \epsilon}\right)^{-1}\right\}$ and ${\vec \phi}=-E\left\{\left(b{\tilde \beta^{\prime T}}\cdot{\tilde \epsilon}\right)^{-1}{\vec N^\prime}\right\}$

The noise of the measurement system can be estimated from the training free model since $b$, ${\tilde \beta^\prime}$ and ${\tilde \epsilon}$ are known. With a set of training samples, the expected noise spectra of a sample $i$ can be obtained from $E\{\ket{n_i}\}=\ket{A_i}-{\tilde \Lambda}^{-1}\cdot{\tilde C}$ and the expected noise power is $E\{\bra{n_i(\lambda)}\ket{n_i(\lambda)}\}$.

\section{Results and Discussions}
\label{sec:result}

In this section, we implemented the aforementioned algorithms and characterized them with the three groups of datasets described in Section~\ref{Sec:Dataset}. For comparison, we also implemented the commonly used  PLSR and XGBoost quantification models.

\subsection{Minimum number of functional and non-functional principal components}

Our first task is to verify that the minimum number of principal components equals the number of independent gas components in the mixture set. Fig.~\ref{fig:min_no_pc}a displays the percentage of the individual (IEV, dashed lines to the left y-axis) and cumulative (CEV, solid lines to the right y-axis) explained variance  vs. the number of top fPCs ranked by their IEV. Here, the fPCs are extracted from Group I training datasets with SNR of 10~dB displayed in cyan, 20~dB in red, 30~dB in green, and 40~dB in blue colors. The black dashed line is positioned at the 9th component. As shown, at the highest SNR of 40~dB where noises are negligible, the top 9 fPCs explain $99.9\%$ of total variance while none of the rest PCs explain more than $1.6{\times}10^{-4}\%$ variance. In this case, almost all variances arising from the 9 gases are retained by the top 9 fPCs. Further studies in the following sections show that each fPC nearly uniquely explains a particular gas, and the fPCs are ranked by the overall absorbance of the corresponding gases. At lower SNR of 30~dB where the signal still dominates the noises, the top 9 components explain almost the same amount of variance as that of 40~dB SNR, while the rising variance explained by the rest of the fPCs is due to the noises. At low SNR of 20~dB, however, none of the fPCs explain more than $6.7\%$ of the total variance. Nevertheless, the first 8 IEVs decrease monotonically, while the rest fPCs show almost the same explained variance. This suggests that although noises start to erode the signal, the first 8 fPCs can still explain the top 8 gases with the strongest absorption coefficient while the rest fPCs are overwhelmed with noises, and the 9th gas with the weakest absorption can not be quantified. At SNR of 10~dB, only the first fPC has a slightly larger IEV, while the rest fPCs show similar variance. Under this situation, only the gas with the largest absorption can be detected while the signals from the rest gases are immersed under noises. It is also noticed that at all SNRs, the IEV of all fPCs other than the top 9 are flat, suggesting that noises are equally distributed among all fPCs due to their truly random nature. Further, it is evident that at lower SNR, part of the variance from the top 9 fPCs that explain signals gradually leaks to all other fPCs at equal distribution.

To further illustrate, we reconstruct the sample spectra by fPCs and compute the average root mean square error (RMSE) to the original spectra.  Showing as solid lines to the left y-axis of Fig.~\ref{fig:min_no_pc}b, at both SNRs of 30~dB (green) and 40~dB (blue), RMSE decreases monotonically when more fPCs are adopted for reconstruction. After 9 fPCs are included, adding more fPCs does not decrease RMSE as they start to explain the variance of noises. At the low SNR of 10~dB, as expected, almost all fPCs explain the noises. As a result, RMSE remains high regardless of the number of fPCs employed for reconstruction. We further study the correlation between the reconstructed and original spectra by defining a residual correlation coefficient $\Delta_\rho=1-\rho$ where
\begin{equation}
    \rho=\frac{\bra{A_r(\lambda)}\ket{A(\lambda)}}{\sqrt{\bra{A_r(\lambda)}\ket{A_r(\lambda)}\bra{A_(\lambda)}\ket{A(\lambda)}}}
\end{equation}
with $\ket{A_r(\lambda)}$ being the spectrum reconstructed by fPCs. As shown by the dashed lines to the right y-axis of Fig.~\ref{fig:min_no_pc}b, $\Delta_\rho$ shows a similar trend to RMSE. At SNRs of 30~dB and 40~dB, $\Delta_\rho$ quickly decreases till the number of involved fPCs increases to 9 and remains flat afterwards. At SNR of 10~dB and 20~dB, little similarity between the reconstructed and original spectra is observed regardless of the number of fPCs included.

Overall, when the SNR is sufficiently high, the top $K$ fPCs form a solution space that can fully explain the signals from all $K$ gases while the rest fPCs are all explaining noises. With decreasing SNR, noises start to ``leak in'' to the solution space spanned by the $K$ fPCs while the signals from the solution space may also ``leak out'' to the space spanned by the rest noise fPCs. As noises and signals are uncorrelated, one may expect that the ``leaked in'' noises will be equally distributed to each of the $K$ fPCs in solution space, and the fPC that explains the least variance may be first ``immersed'' by the noises, making the prediction of the least absorbing gas impossible. Meanwhile, the ``leaked out'' signal will also be equally distributed to the rest $M_\lambda-K$ fPCs. As $M_\lambda\gg{K}$, including only a few additional fPCs into the solution space will not significantly improve prediction results. Therefore, selecting the number of top fPCs that equals the number of gas components in the mixture is sufficient for spectral quantification.

Further, we apply a similar analysis using PCA to the same datasets and obtain similar results (see details in Supplementary Information Section~S.1). In fact, a plot of RMSE and $R^2$ between the functional and non-functional principal components shows that at sufficient high SNR, the differences of the top 9 functional and non-function principal components are minimal as they all explain gas components while other principal components are uncorrelated since the noises are random in nature.

\subsection{Spectral Quantification}
\label{sec:quanti_result}
\subsubsection{Linear regression model}
\begin{table*}[htb]
	\centering
	\caption{Molar extinction coefficient norm of each gas.\label{tab:peak_molar_extinction_ratio}}
	\begin{tabular}{|c | c | c | c | c | c | c | c | c | c |}
	    \hline
	     Gas   &  \ch{N_2O}  & \ch{CO} & \ch{H_2O} & \ch{NO} & \ch{CH4} &   \ch{HCl} & \ch{HF} & \ch{C2H6} & \ch{HBr} \\
	     \hline
	     $\norm{\epsilon}$ (${M}^{-1}{cm}^{-1}$)& 1,166.4  & 569.1 & 371.2 & 219.7 & 162.0 & 160.7 & 126.9 & 103.1 & 30.5\\
		\hline
	\end{tabular}
\end{table*}

In spectral quantification, we first applied fPCA-LR model to quantify the gas components by adopting a 10-fold testing scheme with a 90:10 splitting ratio of the training and testing data samples. In training, fPCA was first applied to the training samples to obtain the top fPCs whose scores were used to train the fPCA-LR model in Fig.~\ref{fig:MLP}. In the testing process, testing samples' fPC scores were computed using the fPCs from training, and spectral quantification was done through the trained model. Fig.~\ref{fig:prediction_rmse_vs_fpc}a-d shows the prediction RMSE of each gas concentration vs. the number of fPCs included. Here and throughout this article, the gases are ordered according to the molar extinction coefficient norm $\norm{\epsilon_k}=\bra{\epsilon_k}\ket{\epsilon_k}$ (Tab.~\ref{tab:peak_molar_extinction_ratio}) from the highest to the lowest. As expected, at all SNRs, the RMSEs of all gases saturate to the minimum at fPCs fewer or equal to 9. In particular, for \ch{HBr} that has the lowest $\norm{\epsilon_k}$, at 20~dB SNR and for all gases at 10~dB SNR, when noises overwhelm the corresponding gas signals, the RMSE will not decrease at a noticeable amount regardless of the number of additional fPCs incorporated. Therefore, it is evident that the required number of fPCs for spectral quantification equals the number of gases in the mixture.

Further, we investigated the RMSE decrement $(-\Delta{RMSE})$ vs. the number of fPCs. As shown in  Fig.~\ref{fig:prediction_rmse_vs_fpc} (e) to (h), at all SNRs except for 10~dB, all gases see a sudden drop of RMSE with the inclusion of a single fPC, with the exception for \ch{CH4} and \ch{HCl} where significant RMSE drop occurs at both 5th and 6th fPCs. The almost one-to-one relation between the gas and fPC, along with observations described in the following sections and Fig.~S7 in Supplementary Information, justify the simplification of quantification using the direct quantification approach, which will be discussed in the following section.

Using 9 fPCs for fPCA-LR quantification, the average testing RMSE of all gases at SNRs of 10~dB, 20~dB, 30~dB, and 40~dB are displayed in Fig.~\ref{fig:RMSE_all_vs_SNR_result}a-d as blue dash-dotted lines with star markers.  In comparison, we also plot the RMSE results of PLSR (red solid lines with plus markers) and XGBoost (green dashed lines with circle markers). In these two models, the 1,000-pixel absorption spectra were used in full as data inputs. As shown, at the highest SNR of 40~dB, our fPCA-LR significantly outperformed PLSR and XGBoost for almost all gases, suggesting that little information was lost, with even as few as 9 fPCs retained. Due to its smallest absorption, the \ch{HBr} result is in line with the PLSR and XGBoost. At SNR of 30~dB, the results from fPCA-LR still outperform gases with strong absorption, but the differences are smaller. For the gases with lower absorption, such as \ch{HCl}, \ch{HF}, \ch{C2H6}, and \ch{HBr}, all three methods have similar performance. Overall, fPCA-LR outperforms PLSR and XGBoost even though it significantly reduced the input data dimension from 1,000 to 9. At lower SNR of 20~dB, however, the rising noises make the prediction of \ch{HBr} as bad as a random guess (black dashed line). Nevertheless, fPCA-LR with 9 fPCs shows performance similar to the 1,000 dimensions PLSR and XGBoost in this case. The trend continues at 10~dB SNR, where none of the methods can be significantly better than random guess predictions for any gases.
\begin{figure*}[htb]
    \centering
    \begin{subfigure}{0.48\textwidth}
    \includegraphics[width=\textwidth]{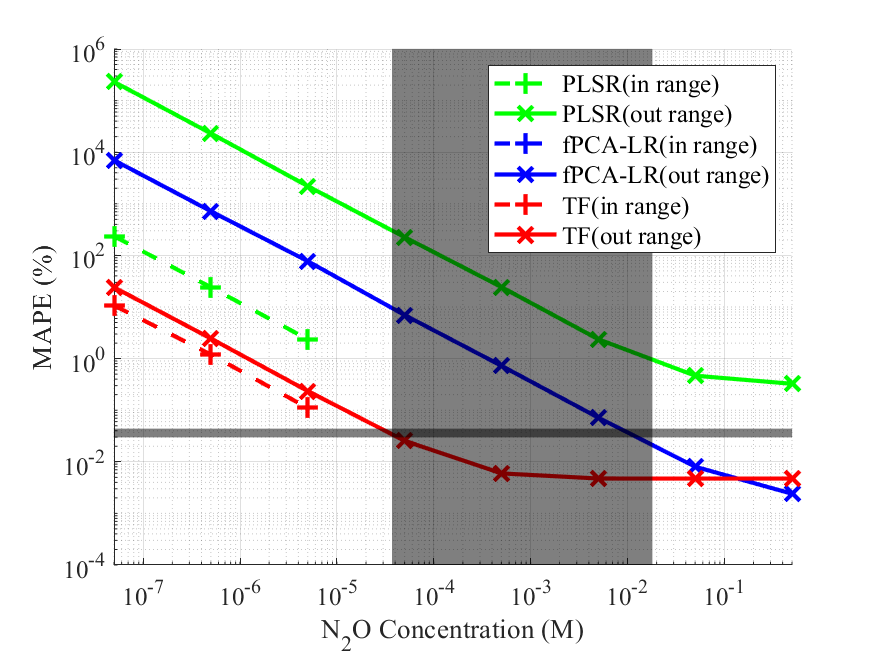}
    \caption{SNR=40~dB}
    \end{subfigure}
\hfill
   \begin{subfigure}{0.48\textwidth}
    \includegraphics[width=\textwidth]{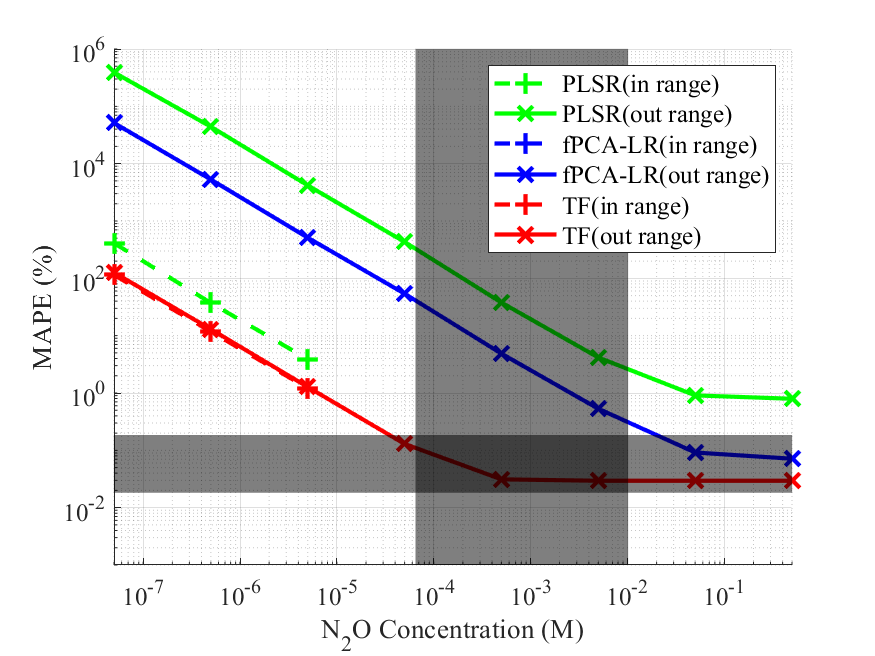}
    \caption{SNR=30~dB}
    \end{subfigure}

    \begin{subfigure}{0.48\textwidth}
    \includegraphics[width=\textwidth]{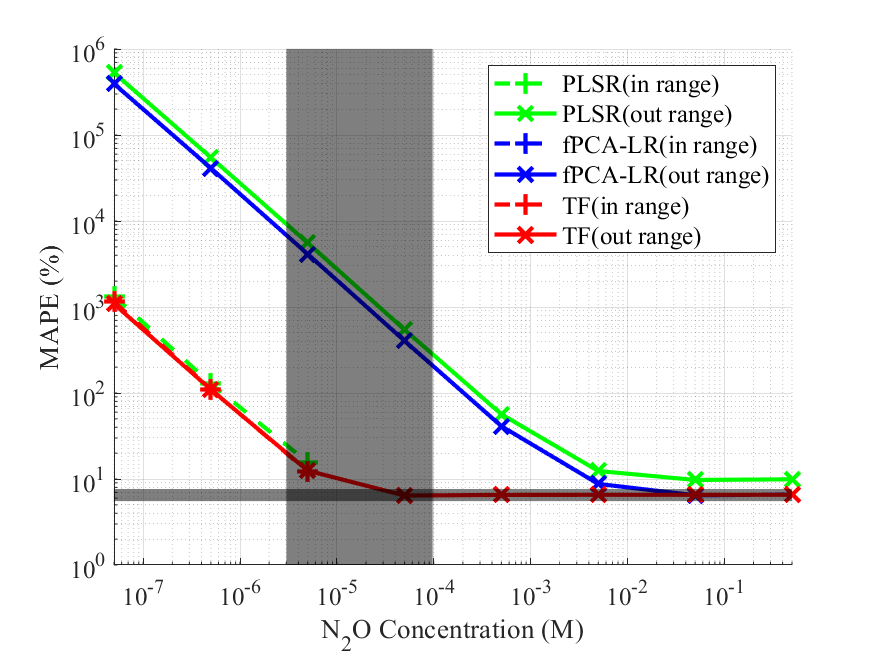}
    \caption{SNR=20~dB}
    \end{subfigure}
\hfill
   \begin{subfigure}{0.48\textwidth}
    \includegraphics[width=\textwidth]{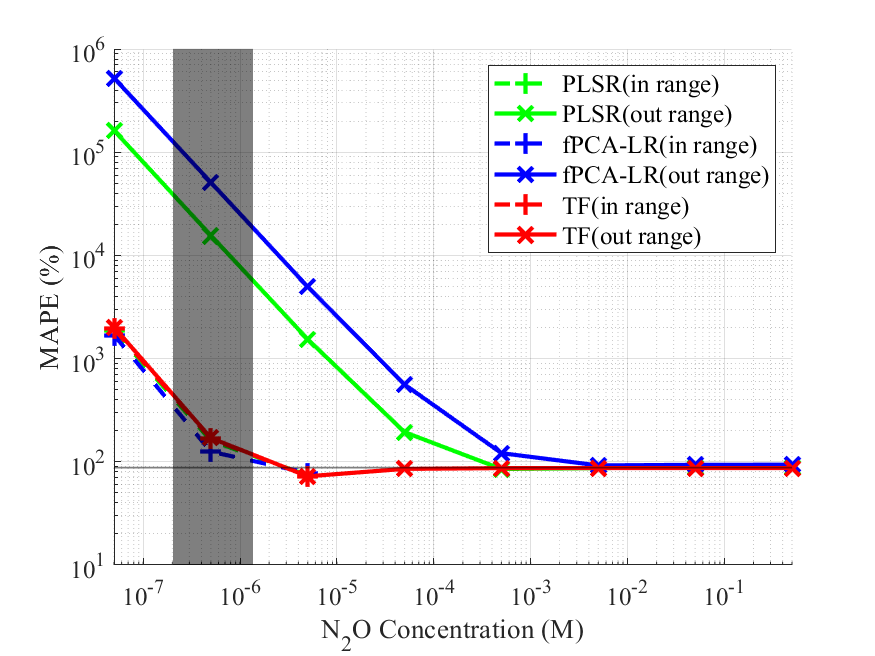}
    \caption{SNR=10~dB}
    \end{subfigure}

    \caption{\label{fig:fig8}The MAPEs of \ch{N2O}  concentration predictions of the samples in datasets that contain concentrations higher than training datasets when the SNRs are (a) 40~dB, (b) 30~dB, (c) 20~dB and (d) 10~dB. The green color represents the PLSR method, the blue represents fPCA-LR and red represents TF method.  The dashed lines with plus markers are the results of the training dataset while the dotted lines with the cross markers are the MAPEs of the dataset contains high concentration. \label{fig:Outrange_prediction}}
\end{figure*}

We further studied the structure of the trained ${\tilde \epsilon}\cdot{\tilde \Lambda}$ at all 4 SNRs and display them in Fig.~\ref{fig:Lambda_heat_map}a-d. It was found that the matrix is highly diagonal at high SNRs. Similar results were observed on the plot of the Pearson correlation coefficients between fPCs and gas molar extinction coefficients (see Fig.~S6 in Supplementary Information). Close studies on the heat maps show that at SNRs of 30~dB and 40~dB, the order of the top fPCs that explain the most variance matches the order of gases ranked by their $\norm{\epsilon_k}$. Further, 7 out of all 9 gases can almost be explained by a single fPC while both \ch{CH4} and \ch{HCl} require 5th and 6th fPC to explain simultaneously. The need for two fPCs to explain the two gases is due to the fact that the norms of their molar extinction coefficients are close ($162.0~{M}^{-1}{cm}^{-1}$ for \ch{CH4} and $160.7~{M}^{-1}{cm}^{-1}$  for \ch{HCl}). Nevertheless, the direct quantification method can be applied to quantification for further reduction of the input data dimension at the cost of lower accuracy. To verify, we plotted the training sample concentrations of \ch{N2O}, \ch{CO}, \ch{H2O}, \ch{NO}, \ch{HF}, \ch{C2H6} and \ch{HBr} vs. their corresponding fPC scores. As shown in Fig.~S2-S5 of Supplementary Information, the linear relation is evident. Therefore, the least square fit can be applied to predict the concentration from the corresponding fPC score directly. Similarly, a two-dimensional linear fit is applied to \ch{CH4} and \ch{HCl}, and we can predict their concentrations in the testing dataset using 5th and 6th fPC scores. It is also worth mentioning that at 20~dB, the weakest absorption gas \ch{HBr} and the 9th fPC decorrelates due to the rising noises. Under this condition, none of the methods can predict its concentration better than random guesses. Fig.~\ref{fig:cross_section_fpc} further compared the molar extinction coefficients and fPC at 30~dB SNR, and both are scaled to their maximum values. As shown in Fig.~\ref{fig:cross_section_fpc}a, the extinction coefficients of \ch{N2O} almost completely coincide with the 1st fPC while Fig.~\ref{fig:cross_section_fpc}b shows the absorption peaks of \ch{CH4} and \ch{HCl} are jointly matched by 5th and 6th fPC. In Fig.~\ref{fig:cross_section_fpc}c, although the peaks of \ch{HBr} are still covered by the 9th fPC, this fPC also starts to explain the rising noises, and partially decorrelates to the \ch{HBr} spectrum. The situation is worse at 10~dB SNR where only \ch{N2O} still slightly correlates to the top fPC. Therefore, \ch{N2O} becomes the only gas whose concentration can be predicted slightly better than a random guess.  Further, we recompute the fPCA-LR by retaining only a few number of leading fPCA coefficients in ${\tilde \Lambda}$. As shown in Fig.~\ref{fig:direct_quantification}a-d, at high SNR of 30 dB and 40 dB, the quantification RMSE of each gas quickly converges with as few as one to two leading matrix elements retained.  Therefore, under an acceptable RMSE, one can reduce the computation complexity by retaining only a few leading matrix elements in ${\tilde \Lambda}$ for quantification.

\subsubsection{Nearly Training free quantification}
As all methods above assume that the molar extinction coefficients of gases are a priori unknown. Here we further tested the TF model proposed in the previous section. In this case, fPCA was applied to the molar extinction coefficients of the 9 gases obtained from the HITRAN database so the sole purpose of the training dataset is to estimate the noise of the dataset, which is unnecessary if noises are negligible. Even with the presence of noticeable noises, the number of training samples required can be orders of magnitude fewer than those for the other algorithms. Fig.~\ref{fig:RMSE_all_vs_SNR_result}a-b illustrates the performance of the TF algorithm. As shown in the pink dotted line with circle markers, the RMSE of our TF model is on par with fPCA-LR, both of which are significantly better than conventional PLSR and XGBoost. Further, we compared the RMSE vs. the number of training samples at all SNRs in Fig.~\ref{fig:RMSE_all_vs_SNR_result}e-h. As shown, at the highest SNR of 40~dB, our TF model (pink dotted line with circle markers) reduces RMSE to lower than 0.1~${\mu}$M with training samples as few as 10. In contrast, fPCA-LR (blue dashed-dotted line) requires around 100 training samples to train the model to similar RMSE, while PLSR (red solid line) and XGBoost (green dashed line) do not reach the same accuracy even while using all 90,000 training samples. This trend maintains at SNRs of 30~dB, 20~dB, and 10~dB where within 100 samples, our TF reduces the RMSE to a level that neither of the other three models can reach with as many as 90,000 samples used for training.

\subsection{Out-of-range spectral quantification}
\label{sec:outlier_concen}
In this subsection, we demonstrate the capacity of the TF model to predict the sample concentrations that are larger than those in the training dataset. In this case, we trained all models with samples of concentrations between 100~pM and 10~$\mu$M (90\% of Group II dataset) and used the trained models to predict sample concentrations in two cases:  a) samples whose constituent concentrations are all in-range with those in the training dataset (the remaining 10\% of the Group II dataset) and b) datasets that have some samples whose constituent concentrations are out-of-range  (10~pM to 1~M, Group III dataset). Fig.~\ref{fig:Outrange_prediction} displays the concentration mean absolute percentage errors (MAPEs) of in-range samples (dashed lines with plus markers) and out-of-range (dotted lines with cross markers) while the green, blue and red colors represent PLSR, fPCA-LR and TF models respectively. Here, XGBoost is not included as it fails to predict out-of-range samples. In this plot, we only show the results of \ch{N2O} while the concentrations of all gases can be found in Figs.~S4-S7 of Supplementary Information. As shown, the in-range sample predictions agree with the results of Fig.~\ref{fig:RMSE_all_vs_SNR_result} where the TF model yields the lowest MAPE seconded by fPCA-LR. Among all methods at all SNRs, PLSR always yields higher MAPE. As expected, for in-range testing, MAPEs decrease monotonically with increasing \ch{N2O} concentration. The performance ranking remains unchanged for out-of-range sample testing as shown in the same figure.  It is observed that for PLSR and fPCA-LR, within the same concentration range, the out-of-range predictions yield higher MAPE than in-range samples. This is because both models learned the noise statistics from the training dataset. When testing with out-of-range samples, even if the sample \ch{N2O} concentration is in the range of the training dataset, the concentrations of other gases in the sample may contribute to ambient noises higher than those in training samples. Therefore, the MAPE becomes larger. On the other hand, the TF model obtains fPCs from the single gas spectra directly, and in concentration prediction, other gases do not contribute to the ambient noises. Therefore, the samples whose \ch{N2O} concentrations are in the range of the training samples yield the same MAPE.

Another intriguing observation is that for the TF model, at low concentrations, the MAPE is inversely proportional to the \ch{N2O} concentration, while at concentrations larger than a threshold $C_{k,th}$, the MAPE saturates to $\gamma_k$.  Detailed analysis in Supplementary Information Section S.3 supports the observation and found
\begin{equation}
\begin{array}{rl}
    \gamma_k=&1-bE\left\{\frac{1}{b_k}\right\}\\
      c_{k,i}^{th}=&\left|\frac{E\left\{\frac{1}{b_k}\right\}\sum_p\Lambda_{k,p}N_{p,i}^\prime-E\left\{N_{k,i}^{''}\right\}}{bE\left\{\frac{1}{b_k}\right\}-1}\right|
    \end{array}
\end{equation}
Fig.~\ref{fig:Outrange_prediction} further plots $\gamma_k\pm{3\sigma_\gamma}$ as the horizontal gray box top and bottom boundaries while the vertical gray box represents the area where the middle 60\% of samples have their $C_{k,i}^{th}$ values fall within it. Here, $\sigma_{\gamma,k}$ is the standard deviation of the 10 $\gamma_k$ values, each obtained from one of the 10-fold testing. As shown, the theory matches the results well. Similar behavior can also be found in the other two models with a larger threshold which may be found analytically through similar approaches.

\section{Conclusions}

In summary, we observed that the essential principal components necessary for quantification in the spectral analysis were equal to the number of independent chemical components included in the sample dataset. Further, we developed fPCA-LR and TF models for spectral quantification. Both models outperformed conventional PLSR and XGBoost at orders of magnitude lower data dimension and the TF model performed the best. We also demonstrated that the fPCA-LR could be further simplified to a direction quantification model due to the near one-to-one relation between principal components and the chemical components in the samples. With the TF model, out-of-range samples can be predicted accurately. The proposed models provided powerful tools for on-site spectral quantification because of their smaller model size, higher accuracy and their requirement for fewer computation resources. In particular, the TF model required few to none training samples, which is essential for spectral quantification as large quantity of training samples is difficult to obtain in practice.

\ifCLASSOPTIONcompsoc
  \section*{Acknowledgments}
\else
  \section*{Acknowledgment}
\fi
The research was funded by the Natural Sciences and Engineering Research Council of Canada (NSERC) Discovery Grants: Grant No. RGPIN‐2017‐04722 (SY and XZ), Grant No. RGPIN-2020-05938 (YB and TL), Grant No. RGPIN‐2021–03530 (XL), Canada Research Chair Grant No. \#950-231363 (XZ), and Threat Reduction Agency (DTRA) Thrust Area 7, Topic G18 (Grant No. GRANT12500317) (YB and TL). This research was enabled in part by support provided by WestGrid (www.westgrid.ca), digital research alliance of Canada (www.alliancecan.ca) and Compute Canada (www.computecanada.ca).

\ifCLASSOPTIONcaptionsoff
  \newpage
\fi


\bibliographystyle{IEEEtran}
\bibliography{ref.bib}

%

\begin{IEEEbiography}[{\includegraphics[width=1in,height=1.25in,clip,keepaspectratio]{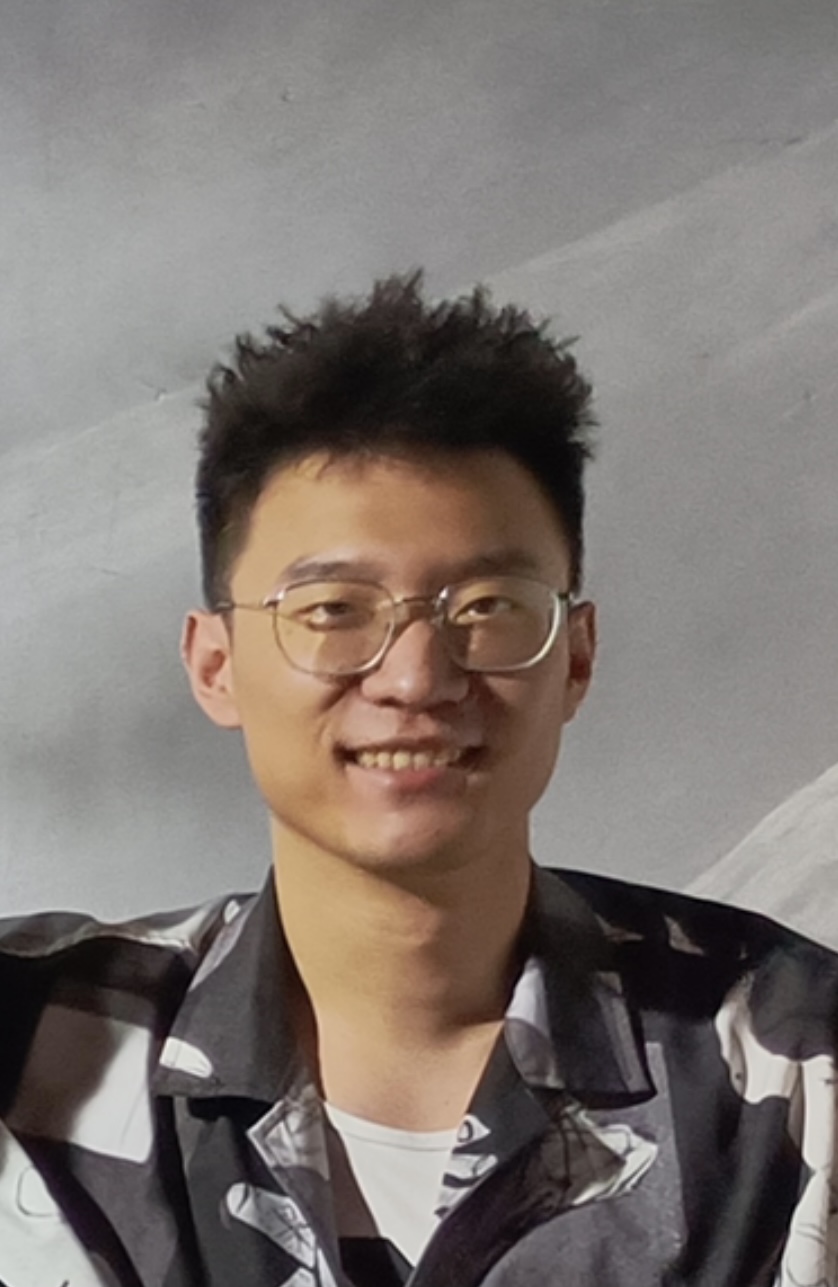}}]{Yifeng Bie}
received the M.A.Sc. degree (2021) and is currently pursing the Ph.D degree in engineering with the University of Victoria. His research interests include machine learning and its application in spectral analysis and numerical modeling. He can be reached at  yifengbie\@uvic.ca
\end{IEEEbiography}

\begin{IEEEbiography}[{\includegraphics[width=1in,height=1.25in,clip,keepaspectratio]{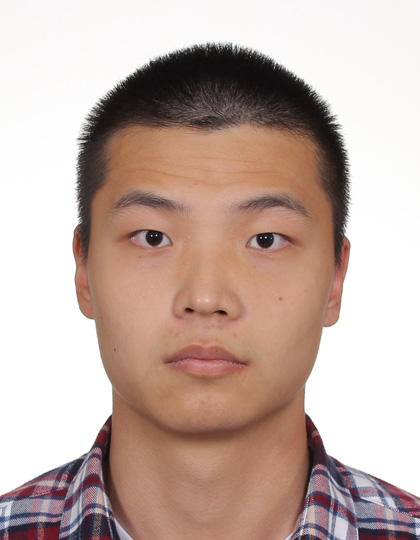}}]{Shuai You} is a fourth-year Ph.D. student in Mathematics and Statistics at the University of Victoria, studying under Professor Xuekui Zhang. Before becoming a graduate student, he completed his B.Sc. in Statistics and a minor in Psychology at the University of British Columbia. His work has been focused on the development of methodologies and software for analyzing data of longitudinal type, which will update and extend his supervisor’s novel methods on health data to answer real-world questions in clinical and medical research. He has also been a teaching assistant at both universities. He can be contacted at shuaiyou\@uvic.ca.
\end{IEEEbiography}

\begin{IEEEbiography}[{\includegraphics[width=1in,height=1.25in,clip,keepaspectratio]{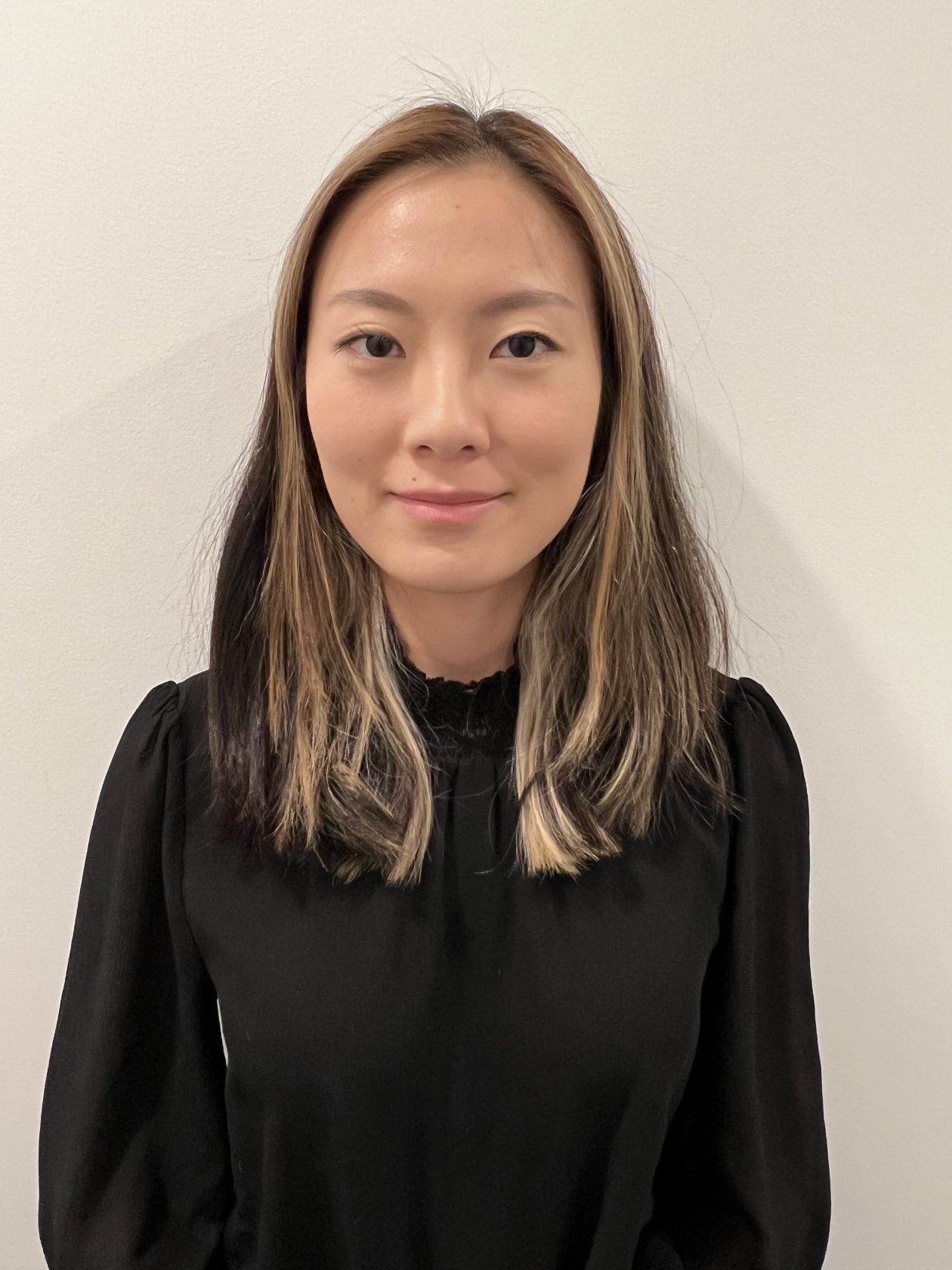}}]{Xinrui Li} is a Data Analyst II at AXYS Technologies Inc., focused on floating LiDAR technology and wind assessment algorithms. She completed her B.Eng. degree in Biomedical Engineering at the University of Victoria. She enjoyed performing hands-on experiments at research laboratories, developing solutions in Python, and breaking down complicated information into reader-friendly stories. She can be contacted at lixinrui97\@gmail.com.
\end{IEEEbiography}

\begin{IEEEbiography}
[{\includegraphics[width=1in,height=1.25in,clip,keepaspectratio]{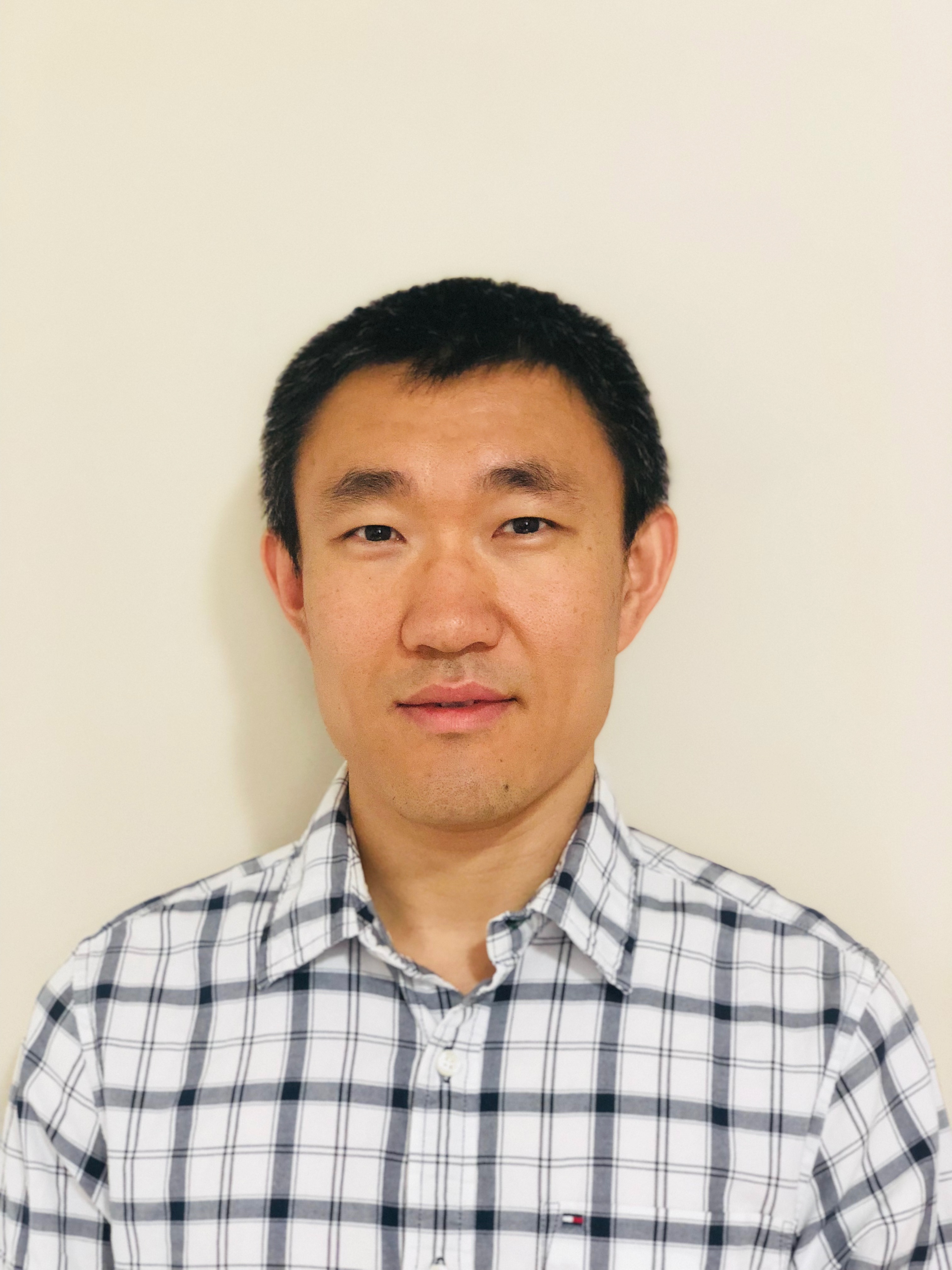}}]{Xuekui Zhang} is an assistant professor in Mathematics and Statistics at the University of Victoria. He is also a Tier 2 Canada Research Chair in Biostatistics and Bioinformatics, and is a Michael Smith Health Research BC Scholar. His research interest is machine learning and its applications in solving real-world big data problems in various areas, such as medicine, agriculture, environment, and economics.
\end{IEEEbiography}

\begin{IEEEbiography}[{\includegraphics[width=1in,height=1.25in,clip,keepaspectratio]{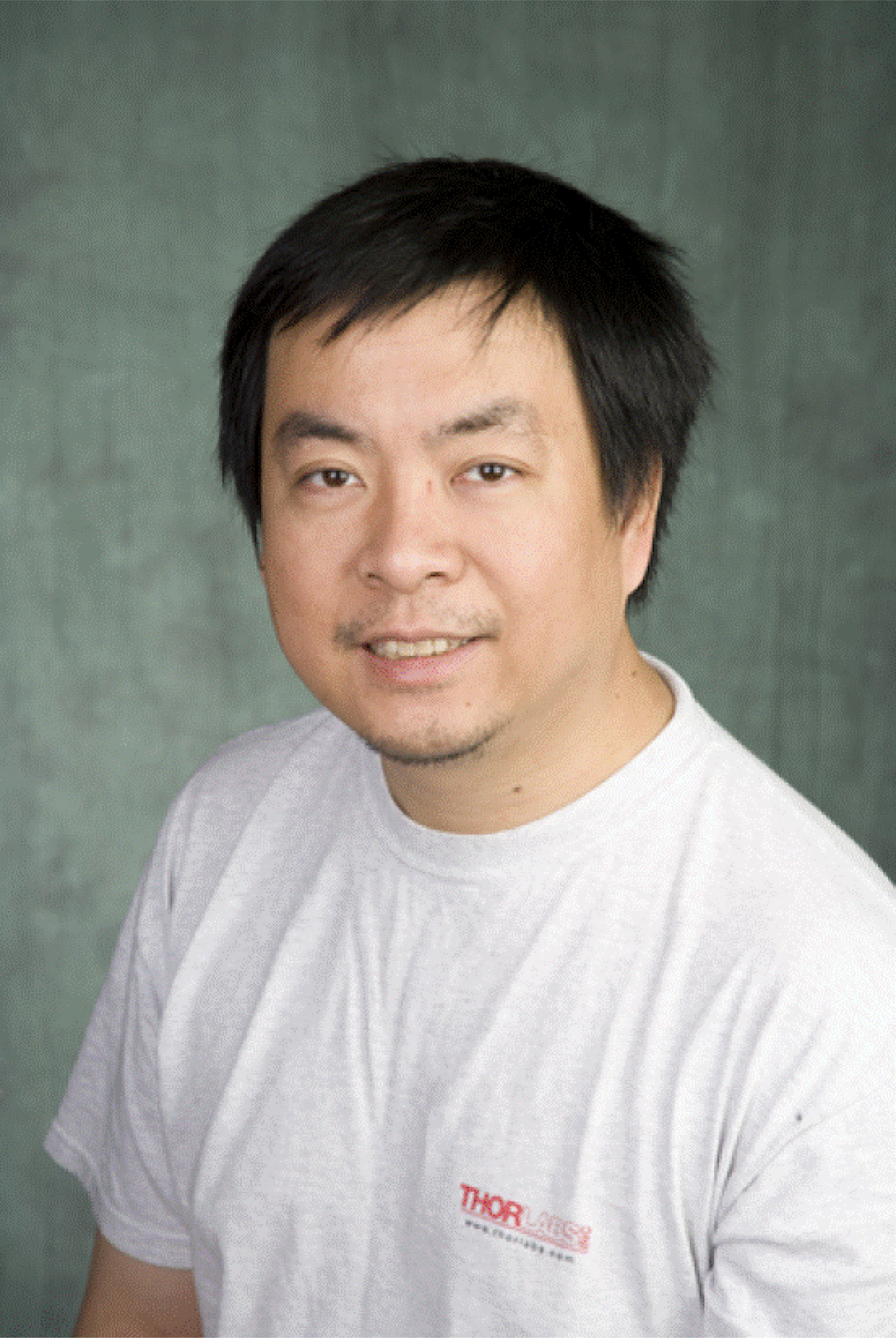}}]{Tao Lu} is an associate professor at Dept. ECE, University of Victoria. He obtained his PhD in 2005 at Dept. Applied Physics, University of Waterloo and has worked in industry with various companies including Nortel Networks, Kymata Canada, Peleton, etc., on optical communications. Before joining the University of Victoria, he was a Post-Doctoral Fellow with the Department of Applied Physics, California Institute of Technology, from 2006 to 2008. His research interests include optical microcavities and their applications to ultra narrow linewidth laser source, and bio-nano-photonics. He is currently extending his research on machine learning algorithms with applications to spectral analysis, Internet of Things, and indoor localization.
\end{IEEEbiography}

\end{document}